\documentclass{article}
\usepackage{ijcai25}

\usepackage{times}
\usepackage{soul}
\usepackage[hidelinks]{hyperref}
\usepackage[small]{caption}
\usepackage[switch]{lineno}

\usepackage{wrapfig}
\usepackage[utf8]{inputenc} 
\usepackage[T1]{fontenc}    
\usepackage{url}            
\usepackage{booktabs}       
\usepackage{amsfonts}       
\usepackage{nicefrac}       
\usepackage{microtype}      
\usepackage{xcolor}         
\usepackage{amsmath}
\usepackage{graphicx}  
\usepackage{float}  
\usepackage{subfigure}  
\usepackage{svg}
\usepackage{algorithm}
\usepackage{algorithmic}
\usepackage{multirow}
\usepackage{enumerate}
\usepackage{amsthm}

\newtheorem{theorem}{Theorem}

\usepackage{multicol}

\title{Variational OOD State Correction for Offline Reinforcement Learning}

\author{
Ke Jiang$^{1,2}$
\and
Wen Jiang$^1$
\And
Xiaoyang Tan$^{1}$
\affiliations
$^1$College of Computer Science and Technology, Nanjing University of Aeronautics and Astronautics, MIIT Key Laboratory of Pattern Analysis and Machine Intelligence\\
\emails
\{ke\_jiang, darren.jum, x.tan\}@nuaa.edu.cn
}

\begin{document}

\maketitle

\begin{abstract}

The performance of Offline reinforcement learning is significantly impacted by the issue of \textit{state distributional shift}, and out-of-distribution (OOD) state correction is a popular approach to address this problem.  In this paper, we propose a novel method named Density-Aware Safety Perception (DASP) for OOD state correction. Specifically, our method encourages the agent to prioritize actions that lead to outcomes with higher data density, thereby promoting its operation within or the return to in-distribution (safe) regions. To achieve this, we optimize the objective within a variational framework that concurrently considers both the potential outcomes of decision-making and their density, thus providing crucial contextual information for safe decision-making. Finally, we validate the effectiveness and feasibility of our proposed method through extensive experimental evaluations on the offline MuJoCo and AntMaze suites.


\end{abstract}

\section{Introduction}

Deep reinforcement learning (RL) has achieved significant success in various domains, including robotics tasks in simulation~\cite{mnih2015human,peng2017deeploco}, game playing~\cite{silver2017mastering}, and large language models~\cite{achiam2023gpt,touvron2023llama}. However, its broader application is constrained by the challenges of interacting with real-world environments, which can be costly or risky~\cite{garcia2015comprehensive}. Offline reinforcement learning addresses these challenges by enabling agents to learn from fixed datasets collected by behavior policies~\cite{zhangzhe}, thereby avoiding high-risk interactions~\cite{lange2012batch}.

Despite this, deploying an online RL framework in an offline setting can significantly hinder the performance of the learned policy. This issue arises from the well-known \textit{distributional shift} problem~\cite{BCQ,cql}, where the TD target may be overestimated for actions with low data density, also known as out-of-distribution (OOD) actions, during training, resulting in extrapolation errors~\cite{pessimism} that degrade the agent's performance.
Previous works, such as Conservative Q-Learning (CQL)\cite{cql}, Bootstrapping Error Accumulation Reduction (BEAR)\cite{bear}, and Supported Policy Optimization (SPOT)\cite{SPOT}, have addressed this problem by suppressing OOD actions through specific regularization techniques. However, these methods primarily focus on avoiding OOD actions while neglecting the issue of \textit{state distributional shift}~\cite{osr,sdc}, which occurs when encountering OOD or low-density states during test, leading to cumulative errors and task failure, i.e., the phenomenon of State deviation.


\begin{figure}[t]
\centering
\setlength{\abovecaptionskip}{0.cm}
\includegraphics[width=0.85\linewidth]{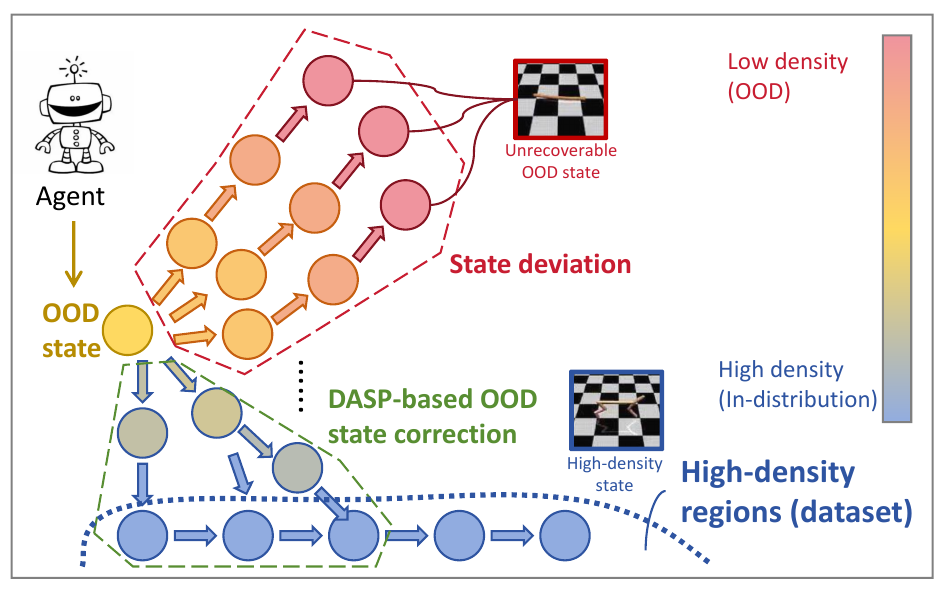}
\caption{The basic idea behind the proposed DASP-based OOD state correction - guiding the agent from OOD states (low-density) to the high density regions according to the dataset.}
\label{fig:framework}
\end{figure}



By OOD states, we mean these states that experience low visitation frequency by the behavior policy. In other words, OOD states exhibit lower density compared to in-distribution states based on the offline dataset. From this perspective, as is shown in Figure \ref{fig:framework}, OOD state correction can be viewed as a process that guides the agent to transition from low-density states to high-density states, ensuring that decision-making is supported by sufficient data and thereby maintaining safety. Such density-based safety requirement is common in online control~\cite{kang2022lyapunov}, but to the best of our knowledge, it has yet to be applied to OOD state correction in offline RL.


In this paper, we introduce a novel method called Density-Aware Safety Perception (DASP) to realize OOD state correction, hence dealing with the problem of \textit{state distributional shift}. 
The basic idea is to guide OOD state correction with an additional reward mechanism based on density optimization. For this purpose, inspired by the likelihood improvement mechanism commonly used in the deep generative model (e.g., diffusion model~\cite{janner2022planning}), we propose a novel offline RL objective that encourages the new policy to prefer to choose those actions that lead to higher data density, besides obtaining higher return. Specifically, we optimize the objective within a variational framework, where DASP predicts the density based on the joint features of the inputted state-action pairs and their potential outcomes. This allows DASP to directly predict one-step forward features and estimate their density to assess the contextual safety of current decision-making, thereby guiding OOD state correction during policy optimization. In practical implementation, our method utilizes a modular algorithmic design, requiring only minor modifications to standard off-policy algorithms to be effective. Our experiments show that the proposed method outperforms several closely related state-of-the-art (SOTA) methods in offline MuJoCo control and AntMaze suites across various settings.

In what follows, after an introduction and a review of related works, Section \ref{sec:preliminary} provides a brief overview of the preliminary knowledge on action constraint methods and consequence-driven methods in offline RL. Section \ref{sec:method} details the DASP method with variational inference and implementation details. Experimental results are presented in Section \ref{sec:exper} to evaluate the effectiveness of the proposed methods under various settings. Finally, the paper concludes with a summary.

\section{Related Works}

\paragraph{Offline reinforcement learning.} The most significant issue in offline RL is balancing conservatism with performance of the learned policy. The Conservative Q-Learning (CQL)~\cite{cql} and Bootstrapping Error Accumulation Reduction (BEAR)~\cite{bear} methods regulate the divergence within a relaxation factor of the new policy. Supported Policy Optimization (SPOT)~\cite{SPOT} takes a different approach by explicitly estimating the behavior policy’s density using a high-capacity Conditional VAE (CVAE)~\cite{cvae} architecture. The most recent advancement in this field is Constrained Policy optimization with Explicit Behavior density (CPED)~\cite{cped}, which utilizes a flow-GAN model to estimate the density of behavior policy more accurately. However, all these methods are to be overly restrictive and lacks robustness and generalization ability, especially at those OOD or unseen states.

\paragraph{OOD state correction.}
OOD state correction methods, also known as state recovery methods, like State Deviation Correction (SDC)~\cite{sdc} align the transitioned distributions of the new policy and the behavior policy, forming a robust transition to avoid the OOD consequences. To further avoid the explicit estimation of consequences in high-dimensional state space, Out-of-sample Situation Recovery (OSR)~\cite{osr} introduces an inverse dynamics model (IDM)~\cite{markovrep} to consider the consequential knowledge in an implicit way when decision making. However, such methods may limit their ability to generalize effectively. State Correction and OOD Action Suppression (SCAS)~\cite{scas} achieves value-aware OOD state correction by state value function and consequence prediction, i.e., aligning high-value transitions of the new policy. However, this method relies on the dynamic model accurately estimating the next state in the transition, which is particularly disadvantageous in the case of stochastic dynamics.


\section{Preliminaries}\label{sec:preliminary}

Reinforcement learning is commonly framed as a Markov Decision Process (MDP), denoted by the tuple $(S,A,P,R,\gamma, \rho_0)$. In this representation, $S$ signifies the state space, $A$ indicates the action space, $P$ is the transition probability matrix, $R$ represents the reward function, $\gamma$ is the discount factor, and $\rho_0$ is the initial state distribution. A policy $\pi:S\rightarrow A$ is established to make decisions during interactions with the environment.

Typically, the Q-value function is expressed as $Q^\pi(s,a) = (1 - \gamma)\mathbb{E} [\sum_{t=0}^\infty\gamma^t R(s_t,\pi(a_t|s_t))|s,a]$, which conveys the anticipated cumulative rewards. For ease of reference, the $\gamma$-discounted future state distribution (or stationary state distribution) is expressed as $d^\pi(s) = (1-\gamma)\sum_{t=0}^{\infty}\gamma^t Pr(s_t=s; \pi,\rho_0)$, with $\rho_0$ representing the initial state distribution and $(1-\gamma)$ acting as the normalization factor.

In an offline context, Q-Learning~\cite{ql} derives a Q-value function $\hat{Q}(s,a)$ and a policy $\pi$ from a dataset $\mathcal{D}$ that is gathered via a behavior policy $\pi_\beta$. This dataset comprises quadruples $(s,a,r,s')\sim d^{\pi_\beta}(s)\pi_\beta(a|s)P(r|s,a)P(s'|s,a)$. The goal is to minimize the Bellman error across the offline dataset~\cite{ql}, employing exact or approximate maximization techniques, such as CEM~\cite{cem}, to retrieve the greedy policy as follows:
\begin{align}
    \min_Q & \mathbb E_{(s,a,r,s')\sim\mathcal D} [r + \gamma\mathbb E_{a'\sim\pi(\cdot|s')}Q(s',a') - Q(s,a) ]^2
    \label{eq:qlearning}\\
    &\quad\quad\quad\quad\quad \max\limits_\pi \mathbb E_{s\sim\mathcal D}\mathbb E_{a\sim\pi(\cdot|s)}[Q(s,a)].\label{eq:qlearning_pi}
\end{align}


\paragraph{OOD State Correction.}
OOD state correction, also known as State recovery, based offline RL methods, such as SDC~\cite{sdc}, OSR~\cite{osr} and SCAS~\cite{scas}, have demonstrate their advantage in developing reliable and robust agents. The basic idea of such methods is to train a policy choosing actions whose state visitation frequency is as closer to that of the behavior policy as possible. It could be represented as follows,
\begin{align}
    \min\limits_\pi \mathbb E_{s\sim \mathcal D}Dis\big(P(\cdot|s, \pi_\beta(\cdot|s)), P(\cdot|s,\pi(\cdot|s))\big)
\end{align}
where $P$ is the dynamics model, and $Dis$ is some kind of distance measure, which is Maximum Mean Discrepancy (MMD) in~\cite{sdc} while Kullback-Leibler (KL) Divergence in~\cite{osr,scas}.

\section{The Method} \label{sec:method}
In this section, we provide a detailed description of the proposed density-aware safety perception framework, termed DASP, to address the issue of \textit{state distributional shift} in offline reinforcement learning. 
\subsection{The Motivation}
Out-of-distribution (OOD) states are defined as those with low density in the dataset, so the aim of OOD state correction is to guide the agent back to high-density regions, thereby ensuring that decision-making is supported by sufficient data. Intuitively, when aiming to identify the high - density regions of offline data, the approach is to leverage the $s_{t}$ distribution information inherent in the dataset. Specifically, techniques such as the diffusion model~\cite{janner2022planning} or score matching~\cite{hyvarinen2005estimation} can be employed to determine the direction in which the $s_{t}$  likelihood experiences an increase, e.g., using a neural network to predict the vector of the score function for a given query state. Subsequently, during deployment, preference is given to the directions that exhibit a high degree of consistency with the likelihood - increasing direction estimated by the score function network. In essence, the action chosen by the agent is a weighted synthesis of two key elements: 1) actions associated with a relatively large reward; 2) actions whose resulting effects align with the direction of the score function.

Nevertheless, a notable limitation of the aforementioned straightforward solution lies in its ignorance of the knowledge context of offline reinforcement learning, failing to account for the impact of factors such as the behavior policy and the environment model during the modeling procedure. In light of this, this paper puts forward a more integrated objective function (Eq.(\ref{eq:oodcorrection1})), as presented in the next section. 

\subsection{Density-Aware Safety Perception}
Given a state \( s \), we first formulate the objective for OOD state correction as follows:
\begin{align}
    \max_{\pi} \mathbb E_{a\sim\pi(\cdot|s),s'\sim P(\cdot|s,a)}\log d^{\pi_\beta}(s')\label{eq:oodcorrection1}
\end{align}
where $P(\cdot|s,a)$ represents the dynamics of the environment, and $d^{\pi_\beta}$ is the stationary state distribution of the behavior policy $\pi_\beta$. The objective in Eq. (\ref{eq:oodcorrection1}) is referred to as Density-Aware Safety Perception (DASP), which evaluates the safety of the input state-action pairs based on the data density of their consequences. We then utilize DASP as a regularization term in policy optimization to prioritize actions that lead the agent toward regions of higher density, thus satisfying safety requirements.

In OOD state correction objective in Eq.(\ref{eq:oodcorrection1}), the $P(\cdot|s,a)$ and $d^{\pi_\beta}$ are two complicated distributions that are hard to estimate explicitly.
Therefore, we implicitly estimate them or their lower bound with the framework of variational inference.
First, we approximate the $d^{\pi_\beta}$ via maximum likelihood estimation, i.e.,
\begin{align}
    d^{\pi_\beta}\approx & \arg\max_{d}\mathbb E_{(s,a,s')\sim\mathcal{D}}\log d(s')\\
    & = \arg\max_{d}\mathbb E_{(s,a)\sim\mathcal{D}, s'\sim P(s'|s,a)}\log d(s')
\end{align}
Then we remark that the estimation of one-step forward density, i.e., $\mathbb E_{s'\sim P(s'|s,a)}\log d(s')$, is the core to realize the OOD state correction. Then Theorem \ref{theorem:varia} gives the solution by estimating the lower bound of the term $\mathbb E_{s'\sim P(s'|s,a)}\log d(s')$ by introducing two variational distributions.

\begin{theorem}\label{theorem:varia}
    The term $\mathbb E_{s'\sim P(s'|s,a)}\log d(s')$ could be lower bounded by solving the following optimization problem in the offline setting,
\begin{align}
    \max_{q_1,q_2}&  \mathbb E_{(s,a,s')\sim\mathcal{D}} \bigg[\int dz\cdot q_1(z|s') \log P(s'|z)\nonumber \\
    -K&L(q_2(z|s,a)\| P(z))- KL(q_1(z|s')\|q_2(z|s,a)) \bigg]\label{eq:final_va_obj}
\end{align}
    where $q_1(z|s')$ and $q_2(z|s,a)$ are two variational distributions. $KL(\cdot\|\cdot)$ is the KL-divergence between two distributions. $P(s'|z)$ is the poster distribution.
\end{theorem}

\textit{The proof is found in Appendix \ref{appendix:proof1}.} In Eq.(\ref{eq:final_va_obj}) the first term represents the reconstruction loss of the consequence $s'$; the second term measures the divergence between the encoding distribution $q_2(z|s,a)$ and the prior distribution $P(z)$, which should be minimized; the third term enables the encoder $q_2$ to directly predict the consequential feature distribution $q_1(z|s')$. This embeds the contextual information into the feature, thereby enabling the decoder to reconstructing the outcome states from either themselves or their previous state-action pairs. The most advantage of this solution is that we can reuse the models to approximate both the dynamics model $P(s'|s,a)$ and the density model $d^{\pi_\beta}(s)$: by the combination of the encoder $q_2(z|s,a)$ and the poster distribution (decoder) $P(s'|z)$, we can predict the consequence of the inputted $(s,a)$; on the other hand, after we have the estimated consequence, we can calculate its density by the variational result in Eq.(\ref{eq:final_va_obj}). The detailed utilization would be discussed in the next section.



Finally, with the objective in Eq.(\ref{eq:final_va_obj}), we can learn the two variational distribution estimators $q_1$ and $q_2$, through which the one-step forward density $\mathbb E_{s'\sim P(\cdot|s,a)}d^{\pi_\beta}(s')$ could be variationally estimated. 
Then, in the next section, we introduce how to utilize this module, also named as DASP, to conduct OOD state correction in an offline manner.

\subsection{DASP-based OOD State Correction}
\begin{figure}[t]
\centering
\setlength{\abovecaptionskip}{0.cm}
\includegraphics[width=1\linewidth]{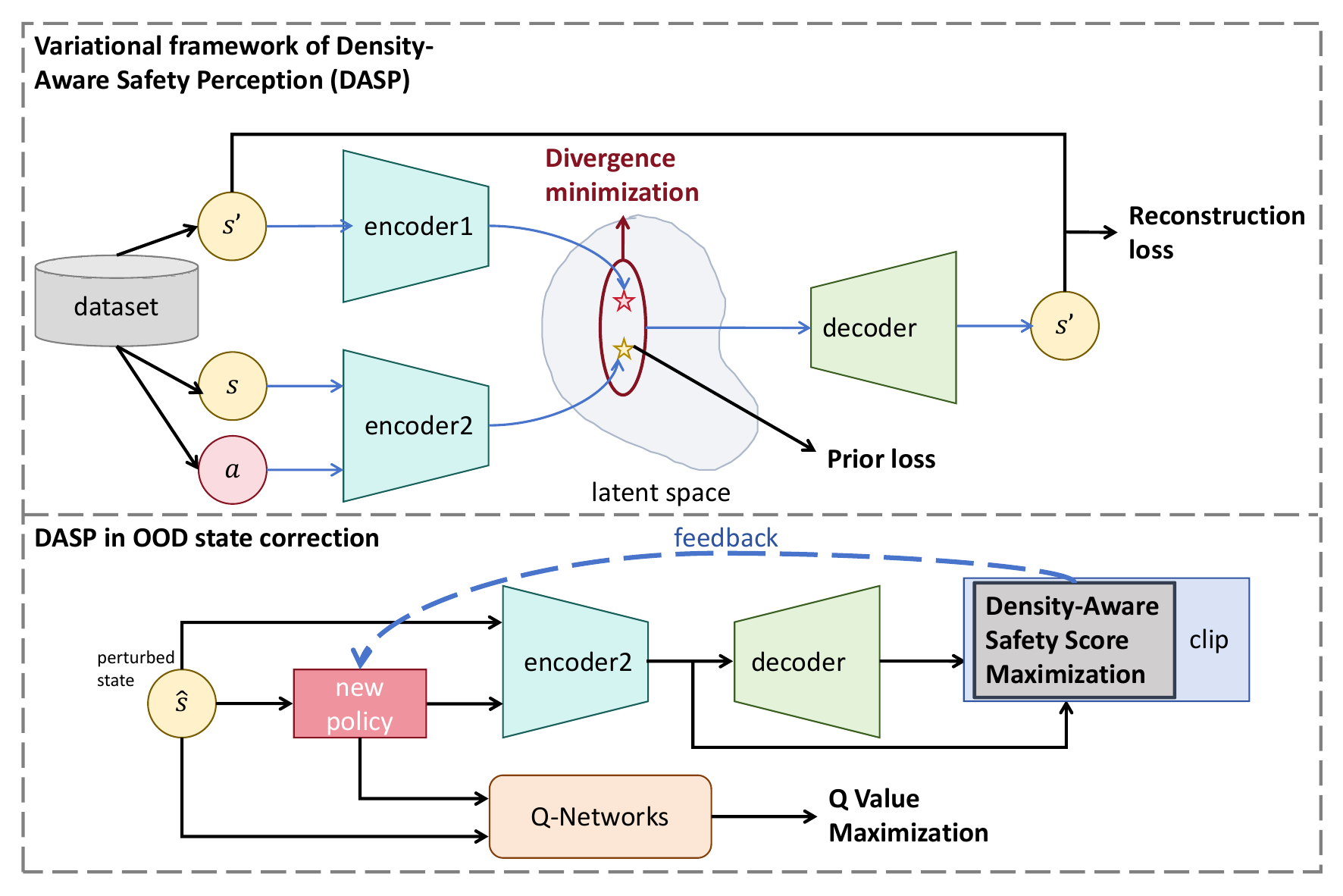}
\caption{The framework of the proposed DASP and its utilization for OOD state correction. In the top figure: the reconstruction loss, prior loss and divergence minimization are the 3 terms in Eq.(\ref{eq:loss_dasp}) respectively. The procedure in the buttom figure represents the policy optimization in Eq.(\ref{eq:actorlossdasp}) .}
\label{fig:implementation}
\end{figure}

First of all, in order to generate OOD states for training, like previous works~\cite{osr,sdc,scas}, we attach Gaussian noise $\mathcal{N}(0,\sigma^2)$ onto the states $s$ from the dataset $\mathcal{D}$, denoted as $\hat{s}$. For OOD state correction in this paper, once the agent entering those OOD states $\hat{s}$, we aim to correct it to restore to safe states with high data density according to the offline dataset. Note that this objective can be reformulated as follows,
\begin{align}
        \max_{\pi} \mathbb E_{s\sim \mathcal{D},\hat{s}\sim\mathbb B_\sigma(s)}\mathbb E_{a\sim\pi(\cdot|\hat{s}), \hat{s}'\sim P(\cdot|\hat{s},a)}\log d^{\pi_\beta}(\hat{s}')
  \label{eq:oodcorrection}
\end{align}
where the $\mathbb B_\sigma(s)$ is a Gaussian perturbation ball with center $s$ and radius $\sigma$. The objective in Eq.(\ref{eq:oodcorrection}) utilizes a one-step forward density module to attach the preference of the actions that could lead to consequences with high data density onto the new policy, hence satisfying the safety requirements for offline RL. Then the practical implementation based on the variational results are as follows,

\paragraph{Parametrization and construction of dynamics model. } Before we handle the policy optimization regularization in Eq.(\ref{eq:oodcorrection}), we need to parameterize the three distribution in Eq.(\ref{eq:final_va_obj}) : the poster distribution $P(s'|z)$ is parameterized with $P_\phi(s'|z)$, which could also be seen as the decoder module; the two variational distribution $q_1(z|s')$ and $q_2(z|s,a)$ are parameterized with $q_{\psi}(z|s')$ and $q_{\theta}(z|s,a)$ (corresponding to two encoders respectively in Figure \ref{fig:implementation}(top)) . In this way, we reformulate the optimization problem by,
\begin{align}
    \theta^*, \psi^*, \phi^*\quad \quad\quad\quad\quad\quad\quad\quad\quad\quad\quad\quad& \nonumber\\
    = \arg\max_{\theta, \psi, \phi} \mathbb E_{(s,a,s')\sim\mathcal{D}} \bigg[\int dz\cdot q_{\psi}(z|s')& \log P_\phi(s'|z)\nonumber \\
    -KL(q_{\theta}(z|s,a)\|& P(z))\nonumber\\
    - KL(q_{\psi}(z|s')\|&q_{\theta}(z|s,a)) \bigg]\label{eq:obj_para}
\end{align}
Then the above optimization could be further solved by methods like in~\cite{vae,burda2015importance}. Specially, all the parameterized distributions are assumes as Gaussian - $q_{\theta}(z|s,a)=\mathcal{N}(\mu_\theta,\sigma_\theta;s,a)$, $q_{\psi}(z|s')=\mathcal{N}(\mu_\psi,\sigma_\psi;s')$ and $P_\phi(s'|z)=\mathcal{N}(\mu_\phi,\sigma_\phi;s')$. Suppose the prior distribution $P(z) = \mathcal{N}(0, I)$, then the above formulation in Eq.(\ref{eq:obj_para}) could be transferred into the loss function as,
\begin{align}
    \mathcal{L}_{dasp} &(s,a,s';\theta,\psi,\phi)= \mathbb E_{z\sim q_{\theta}(z|s,a)}\|\mu_\phi(z) - s'\|^2_2\nonumber\\
     - \frac{1}{2}&\big[\sum_i^K(1+\log(\sigma_{\theta,i}^2) - \mu_{\theta,i}^2 - \sigma_{\theta,i}^2)\big]\nonumber\\
     - \frac{1}{2}&\big[\sum_{i=1}^K(\log\frac{\sigma_{\psi,i}}{\sigma_{\theta,i}}+\frac{(\sigma_{\psi,i})^2+(\mu_{\psi,i} - \mu_{\theta,i})^2}{2(\sigma_{\theta,i})^2})\big]\label{eq:loss_dasp}
\end{align}
where $i$ represents the value of $i^{th}$ dimension of the $K$-dimensional variable.

The forward dynamics model $P(s'|s,a)$ could be estimated by the combination of $P_{\phi^*}(s'|z)$ and $q_{\theta^*}(z|s,a)$. Here the $q_{\theta^*}(z|s,a)$ could be seen as an approximation of $q_{\psi}(z|s')$ due to the minimization of the divergence between the representations generated by these two encoders, hence the combined module could predict the $s'\sim P(s'|s,a)$ from $z\sim q_{\theta^*}(z|s,a)$ with low bias. Then the approximated dynamical model is denoted as $\hat{P}(s'|s,a)$.

\begin{table*}[h]
\centering
\caption{Results of \textbf{DASP(ours)}, CQL, PBRL, SPOT, SVR, EDAC, RORL, SDC , OSR-10 and SCAS on D4RL averaged over 4 seeds. We bold the highest scores in each task.}
\begin{tabular}{ll|lllllllll|l}
\toprule
                             &     & CQL  & PBRL  & SPOT  & SVR   & EDAC  & RORL  & SDC   & OSR-10 & SCAS  & DASP(Ours) \\ \midrule
\multirow{5}{*}{\rotatebox[
origin=c]{90}{halfcheetah}}  & r   & 17.5 & 11.0  & 35.3  & 27.2  & 28.4  & 28.5  & \textbf{36.2}  & 26.7  & 12.2 & 32.4±0.9  \\
                             & m   & 47.0 & 57.9  & 58.4  & 60.5  & 65.9  & 66.8  & 47.1  & 67.1  & 46.6 & \textbf{70.4}±2.9  \\
                             & m-e & 75.6 & 92.3  & 86.9  & 94.2  & 106.3 & 107.8 & 101.3 & 108.7 & 91.7  & \textbf{112.1}±2.0 \\
                             & m-r & 45.5 & 45.1  & 52.2  & 52.5  & 61.3  & 61.9  & 47.3  & 64.7  & 44.0  & \textbf{67.1}±3.9  \\
                             & e   & 96.3 & 92.4  & 97.6  & 96.1  & 106.8 & 105.2 & 106.6 & 106.3 & 106.6 & \textbf{107.4}±1.8 \\ \hline
\multirow{5}{*}{\rotatebox[
origin=c]{90}{hopper}}       & r   & 7.9  & 26.8  & 33.0  & 31.0  & 25.3  & 31.4  & 10.6  & 30.4  & 31.4  & \textbf{33.1}±0.3  \\
                             & m   & 53.0 & 75.3  & 86.0  & 103.5 & 101.6 & 104.8 & 91.3  & 105.5 & 102.5 & \textbf{108.6}±0.9 \\
                             & m-e & 105.6& 110.8 & 99.3  & 111.2 & 110.7 & 112.7 & 112.9 & 113.2 & 109.7 & \textbf{116.0}±6.3 \\
                             & m-r & 88.7 & 100.6 & 100.2 & 103.7 & 101.0 & 102.8 & 48.2  & 103.1 & 101.6 & \textbf{104.1}±1.1 \\
                             & e   & 96.5 & 110.5 & 112.3 & 111.1 & 110.1 & 112.8 & 112.6 & \textbf{113.6} & 112.8 & 113.5±1.0 \\ \hline
\multirow{5}{*}{\rotatebox[
origin=c]{90}{walker2d}}     & r   & 5.1  & 8.1   & 21.6  & 2.2   & 16.6  & 21.4  & 14.3  & 19.7  & 1.4   & \textbf{23.9}±0.8   \\
                             & m   & 73.3 & 89.6  & 86.4  & 92.4  & 92.5  & 102.4 & 81.1  & 102.0 & 82.3  & \textbf{108.6}±2.7  \\
                             & m-e & 107.9& 110.8 & 112.0 & 109.3 & 114.7 & 121.2 & 105.3 & \textbf{123.4} & 108.4 & 123.0±2.6  \\
                             & m-r & 81.8 & 77.7  & 91.6  & 95.6  & 87.1  & 90.4  & 30.3  & 93.8  & 78.1  & \textbf{99.5}±1.7   \\ 
                             & e   & 108.5& 108.3 & 109.7 & 110.0 & 115.1 & \textbf{115.4} & 108.3 & 115.3 & 115.0 & 115.3±1.6  \\ \hline
average                      &     & 67.4 & 74.4  & 78.8  & 80.0  & 82.9  & 85.7  & 70.2  & 86.2  & 76.3  & \textbf{89.0}       \\ 
\midrule
\multirow{5}{*}{\rotatebox[
origin=c]{90}{antmaze}}      & umaze      & 82.6 & -    & 93.5  & -     & -     & \textbf{96.7}  & 81.4  & 89.9  & 90.4  &  94.6±3.2 \\
                             & umaze-div  & 10.2 & -    & 40.7  & -     & -     & \textbf{90.7}  & 49.6  & 74.0  & 63.8  &  65.5±6.1 \\
                             & med-play   & 59.0 & -    & 74.7  & -     & -     & 76.3  & 55.0  & 66.0  & 76.6  &  \textbf{79.0}±4.6 \\
                             & med-div    & 46.6 & -    & 79.1  & -     &  -    & 69.3  & 56.6  & 80.0  & \textbf{80.4}  &  79.6±4.9 \\ 
                             & large-play & 16.4 & -    & 35.3  & -     &  -    & 16.3  & 20.8  & 37.9  & 49.0  &  \textbf{49.3}±8.5 \\ 
                             & large-div  & 3.2  & -    & 36.3  & -     &  -    & 41.0  & 25.8  & 37.9  & \textbf{50.6}  &  43.4±9.3 \\ \hline
average                      &            & 36.3 & -    & 59.9  & -     &  -    & 65.1  & 48.2  & 64.3  & 68.5  &  \textbf{68.6} \\ 
\bottomrule
\end{tabular}
\label{comparison}
\end{table*}

\paragraph{DASP-based actor regularization. }We construct the estimation term\footnote{The validation study for this term is shown in Sec.\ref{sec:validation}.} for the objective in Eq.(\ref{eq:oodcorrection}) based on the variational results and the parameterization. To be specific, the OOD state correction term could be approximated by,
\begin{align}
    \mathcal{R}(\hat{s},a) = \mathbb E_{\hat{s}'\sim \hat{P}(\cdot|\hat{s},a)}f_{\tau}(\mathcal{L}_{dasp}(\hat{s},a,\hat{s}';\theta^*,\psi^*,\phi^*))\label{eq:calculationofscore}
\end{align}
where $(\theta^*,\psi^*,\phi^*)$ is the solution by minimizing the DASP loss $\mathcal{L}_{dasp}$ over the dataset $\mathcal{D}$ and $f_\tau$ is a clip function with threshold $\tau$. The use of the clipping function $f_\tau$ is motivated by our objective, which is not to maximize likelihood but to regularize the agent's visitation to ensure sufficient density, specifically above a specified threshold $\tau$.
Please note that, instead of pretraining the dynamics model separately, the $\hat{P}(s'|s,a)$ is constructed by the modules in $\mathcal{L}_{dasp}$, hence formulating the indicator $\mathcal{R}(\hat{s},a)$ a more compact implementation compared with other methods~\cite{sdc,osr,scas}. The actor loss is,
\begin{align}
    \max\limits_\pi \mathbb E_{s\sim\mathcal D}\Big[\mathbb E_{a\sim\pi(\cdot|s)}&[Q(s,a)]\nonumber\\
    &+ \alpha\cdot\mathbb E_{\hat{s}\sim\mathbb B_\epsilon(s),a\sim\pi(\cdot|\hat{s})}\mathcal{R}(\hat{s},a)\Big]\label{eq:actorlossdasp}
\end{align}
where $\alpha$ is the balance coefficient of the DASP term. Besides, we also utilize a momentum-based optimizer, e.g. Adam, in implementation to avoid the problem of local optimum.


\paragraph{Overall Algorithm. } Figure \ref{fig:implementation} gives the network architecture of the proposed DASP approach, while the whole training algorithm is shown in Algorithm \ref{alg1}. 

\begin{algorithm}[H]
    \caption{DASP-based offline RL framework}
    \label{alg1}
    \textbf{Input}: offline dataset $\mathcal{D}$, maximal update iterations $T$,\\
    \textbf{Parameter}: policy network $\pi$, Q-networks $Q_1,Q_2$, DASP module $\mathcal{R}$,\\
    \textbf{Output}: learnt policy network $\pi$
    \begin{algorithmic}[1] 
    	\STATE Initialize the policy network, Q-networks and the DASP module.
        \STATE Pretrain the DASP module $\mathcal{R}$ according to Eq.(\ref{eq:loss_dasp}).
        \STATE Let $t=0$.
        \WHILE{$t < T$}
        \STATE Sample mini-batch of N samples $(s,a,r,s')$ from $\mathcal{D}$.
        \STATE Perturb $s$ with Gaussian Noise and get $\hat{s}$.
        \STATE Feed $\hat{s}$ into the policy network, get the action $a$ and calculate the DASP score $\mathcal{R}(\hat{s},a)$.
        \STATE Update the Q-networks according to Eq.(\ref{eq:qlearning}),
		\STATE Update the policy network $\pi$ according to Eq.(\ref{eq:actorlossdasp}).

        \ENDWHILE
        \STATE \textbf{return} learnt policy network $\pi$.
    \end{algorithmic}
\end{algorithm}

\begin{table*}[h]
    \caption{Results of RORL, SDC, OSR-10 and DASP in OOSMuJoCo setting on the normalized return and decrease metric averaged over 4 seeds. The noteworthy results are bolded.}
    \label{zaoshengshiyan}
    \centering
    \begin{tabular}{lllllllll}
        \toprule
							& \multicolumn{2}{l}{RORL} & \multicolumn{2}{l}{SDC} & \multicolumn{2}{l}{OSR-10}  & \multicolumn{2}{l}{DASP}\\
Task name                  & score & dec.(\%) & score & dec.(\%) & score & dec.(\%) & score & dec.(\%)  \\
			        \midrule
Halfcheetah-OOS-slight   & 55.3  & 17.2 & 45.1 & \textbf{4.3} &\textbf{59.4} &  11.5      & 58.5±1.2 & 14.8      \\
Halfcheetah-OOS-moderate & 47.6  & 28.7 & 39.8 & \textbf{15.5}  &56.5 & 15.8   & \textbf{56.9±2.2} & 17.2        \\
Halfcheetah-OOS-large    & 35.4 & 47.0 & 34.0 &  27.8  & 50.8 & 24.3        & \textbf{54.6±4.2} & \textbf{20.5}     \\  \hline
Hopper-OOS-slight       & 100.4 & 4.2   & 85.7 & 6.1  &100.8 & 4.5     & \textbf{101.9±0.2} &  \textbf{4.1}      \\
Hopper-OOS-moderate      & 94.4 & 9.9  & 82.9 &  9.2  &98.3 &  \textbf{6.8}   & \textbf{98.5±0.5} &  \textbf{7.3}       \\
Hopper-OOS-large         & 82.1 & 21.7   & 75.5 & 17.3  & \textbf{94.7} &  \textbf{10.2}     & 89.5±2.4 &	15.8\\ \hline
Walker2d-OOS-slight      & 92.9 & \textbf{9.3}  & 71.0 & 12.5  &92.4 & 9.4   & \textbf{93.3±0.7} &  10.4      \\
Walker2d-OOS-moderate    & 86.5 &  15.5  & 69.5 & 14.3  &90.3 & \textbf{11.5}  & \textbf{91.4±1.1} & \textbf{12.2}     \\
Walker2d-OOS-large       & 71.8 & 29.9   & 65.3 &  19.5    &88.6 & \textbf{13.1}   & \textbf{89.1±4.6} &  \textbf{14.4}  \\
        \bottomrule
    \end{tabular}
\end{table*}

\begin{figure*}[h]
    \centering
    \setlength{\abovecaptionskip}{0.cm}
    \includegraphics[width=0.9\linewidth]{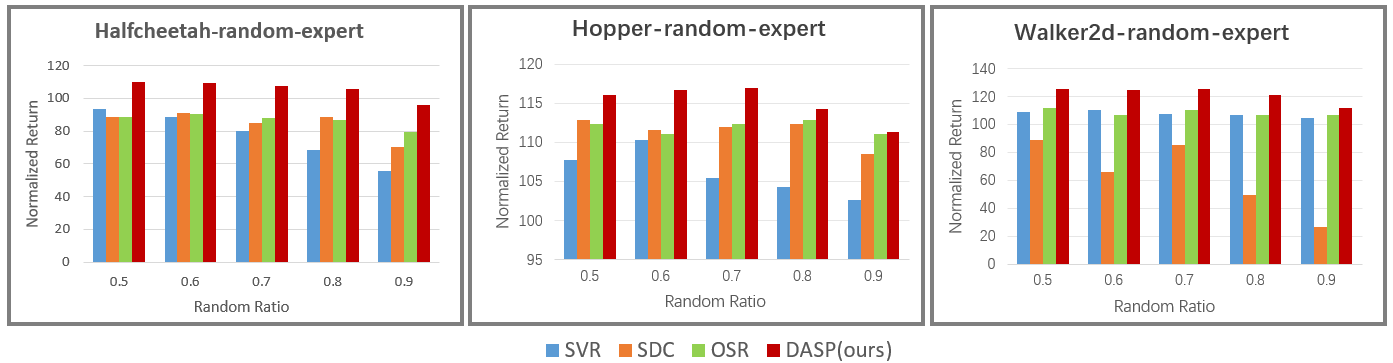}
    \caption{The results on the MuJoCo benchmarks with different levels of sub-optimal data.}
    \label{fig:limitedvaluable}
\end{figure*}

\section{Experiments}\label{sec:exper}

In experiments we answer the following three key questions:
\begin{enumerate}[1)]
\item Does DASP achieve the state-of-the-art performance on standard MuJoCo benchmarks compared to the latest closely related methods?\label{question:1}
\item Is DASP able to recover from out-of-distribution (OOD) states successfully?\label{question:2}
\item Is DASP term robust enough to deal with unfavorable conditions, such as sub-optimal demonstrations or inefficient samples, in practical deployments?\label{question:3}
\end{enumerate} 

Our experimental section is organized as follows: 
First, by fairly comparing the performance of learning policies using traditional methods on standard MuJoCo benchmarks, we verify that the proposed method DASP achieves superior performance among these methods, answering Question \ref{question:1}.
Then, to answer Question \ref{question:2}, we verify the ability of DASP to recover from OOD states using the Out-of-sample MuJoCo (OOSMuJoCo) benchmarks, as described in~\cite{osr}.
Finally, to answer Question \ref{question:3}, we evaluate DASP on benchmarks under the settings of sub-optimal data and inefficient data~\cite{sdc}.
Additionally, we conducted an ablation study and designed an experiment to analysis the validity of the DASP regular term. 
A brief introduction of our code is available in Appendix \ref{code}.

\subsection{Comparisons on Standard Benchmarks}

In this section, we compare the two proposed implementations of our method with several significant methods, including CQL~\cite{cql}, PBRL~\cite{pbrl}, SPOT~\cite{SPOT}, SVR~\cite{svr}, EDAC~\cite{edac}, RORL~\cite{rorl}, SDC~\cite{sdc}, OSR-10~\cite{osr} and SCAS~\cite{scas}, based on the D4RL~\cite{d4rl} dataset in the standard MuJoCo benchmarks and AntMaze tasks. 

\textbf{MuJoCo (D4RL).} The MuJoCo domain have three types of high-dimensional control environments representing different robots in D4RL: Hopper, Halfcheetah and Walker2d, and five kinds of datasets: 'random', 'medium', 'medium-replay', 'medium-expert' and 'expert'. 
The \textbf{AntMaze} domain is a more challenging navigation domain with sparse rewards and multitask data, which contains three types of datasets, namely ‘umaze’, ‘medium’, and ‘large’.

The results is shown in Table \ref{comparison}, where part of the results for the comparative methods are obtained by~\cite{rorl,osr,scas}. On the MuJoCo tasks, we have observed that the performance of all methods experiences a significant decrease when learning from  datasets  such as 'random', 'medium', 'medium-replay', and 'medium-expert', which are collected by sub-optimal behavior policies. This highlights the inherent difficulty in getting rid of the influence on the sub-optimal behavior strategy in practical settings. However, our proposed methods, DASP, consistently outperform other approaches across most benchmarks, particularly surpassing methods that rely on behavior cloning such as CQL, PBRL, and EDAC. Furthermore, DASP achieve state-of-the-art performance in terms of the average score. Additionally, we would like to emphasize that DASP demonstrates significant improvements over the state-of-the-art conservative methods (e.g., SVR and OSR) on the 'medium' and 'medium-replay' datasets. This notable margin can be attributed to DASP’s ability to avoid aligning the transition of the dataset through its flexibility in correcting the consequences. This further underscores the advantages of DASP in effectively handling sub-optimal offline data. In the following section, we will explore DASP’s ability to recover from OOD states. On the AntMaze tasks, DASP outperforms all the methods in total score, and is very close to SOTA method in each item.

\begin{figure*}[h]
\centering
\setlength{\abovecaptionskip}{0.cm}
\includegraphics[width=0.85\linewidth]{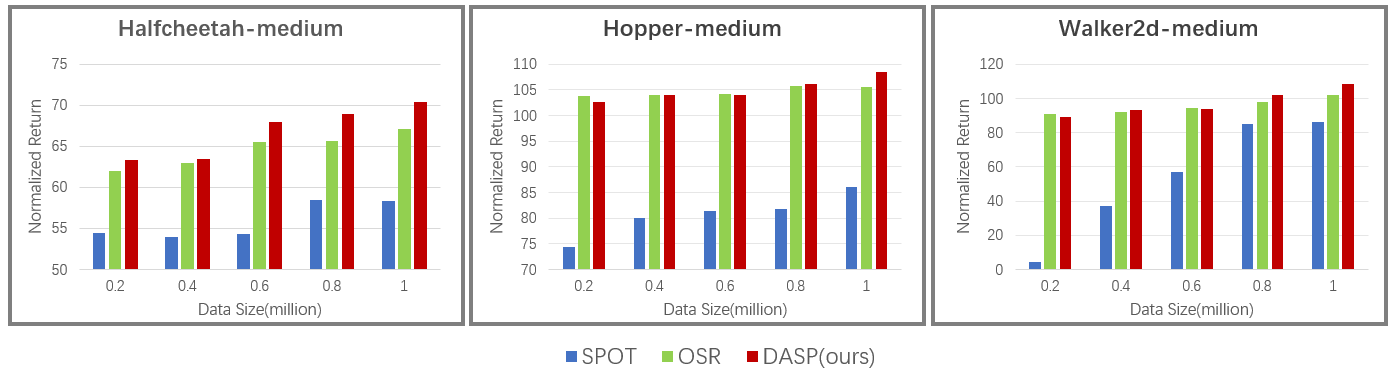}
\caption{The results on three MuJoCo benchmarks with different size of 'medium' datasets.}
\label{fig:differentsize}
\end{figure*}

\subsection{Evaluation on Out-of-sample MuJoCo Setting}
To investigate the agent’s behavior in unseen (OOD) states and assess whether the proposed DASP enables recovery from out-of-sample situations, we introduce the OOSMuJoCo benchmarks from~\cite{osr} and implement other related methods: RORL, SDC, and OSR-10 on ‘medium’ datasets. OOSMuJoCo simulates external forces to push the agent into out-of-sample states in Halfcheetah, Walker2d, and Hopper, with three levels of force: slight, moderate, and large.

Table \ref{zaoshengshiyan} presents the scores and performance decreases of these policies across the 9 OOSMuJoCo benchmarks. The performance decrease is calculated as the percentage reduction in scores from OOSMuJoCo compared to the standard MuJoCo environments shown in Table \ref{comparison}. The results indicate that the proposed DASP outperforms other methods in scores, particularly in the ‘Halfcheetah’ and ‘Walker2d’ benchmarks with larger perturbations, likely due to these benchmarks’ higher sensitivity to OOD situations. Additionally, we note that DASP and OSR-10 exhibit comparable performance decreases across the environments, suggesting that methods incorporating the DASP constraint are at least as robust as OSR-10 and RORL in handling OOD situations. Next, we will explore DASP’s capabilities in sub-optimal demonstrations.

\subsection{Evaluation on Sub-Optimal Datasets}
In this section, we further investigate the feasibility of the proposed DASP on different levels of sub-optimal offline datasets, where ‘expert’ and ‘random’ datasets are mixed in various ratios. This setting is widely used, as seen in~\cite{sdc,svr,osr}. 
In this paper, the proportions of ‘random’ data are 0.5, 0.6, 0.7, 0.8, and 0.9 for ‘Halfcheetah’, ‘Hopper’, and ‘Walker2d’. 

We compare the proposed DASP with SVR~\cite{svr}, OSR~\cite{osr}, and SDC~\cite{sdc}. As shown in Figure \ref{fig:limitedvaluable}, our method outperforms the other three methods across the three control environments in terms of normalized scores. We observed that our proposed method exhibits a significantly lower decrease rate over the ‘Halfcheetah’ benchmark compared to the other two methods as the random ratio increases, which can be attributed to the agent’s heightened sensitivity to the quality of data collection in this environment. Furthermore, when testing on the ‘Hopper’ and ‘Walker2d’ benchmarks, we note that DASP demonstrates the least decrease in performance among all methods when the random ratio reaches 0.9. This highlights the advantage of the implicit implementation in addressing more complex tasks and learning from lower-quality data in practical scenarios. Therefore, we emphasize that our method is better equipped for learning with sub-optimal data and exhibits improved stability and performance across various benchmarks.

\subsection{Evaluation on Data Inefficient Benchmarks}\label{sec:differentsize}
Sub-optimal data can be considered as a form of noisy-labeled data, where certain states ‘$s$’ are associated with sub-optimal (incorrect) labels, denoted as action ‘$a$’. 
Previous studies~\cite{wang2014robust,bootkrajang2012label} have shown that learning performance is significantly influenced by the size of the training data. This motivated us to investigate the performance of different methods under varying sizes of sub-optimal data.

In this section, as depicted in Figure \ref{fig:differentsize}, we compare our proposed DASP method with typical offline RL approaches, namely SPOT and OSR-10, using different sizes of training data (0.2, 0.4, 0.6, 0.8 million). We select the ‘medium’ datasets as the sub-optimal training data. Our observations reveal that the DASP method consistently outperforms the other two methods across all data sizes. 
Notably, both DASP and OSR-10 exhibit superior performance compared to SPOT by a significant margin.
Furthermore, the advantage of DASP over OSR-10 becomes more pronounced as the data size increases. 
These findings demonstrate that the challenges of dealing with OOD states in offline RL would diminish with massive data sizes. However, when the data is insufficient, OOD state correction methods ,including our proposed DASP, exhibit better generalization capabilities.

\subsection{Validity Analysis of DASP Regularization}\label{sec:validation}


In this section, we perform a experiments within the MuJoCo environment to Analysis the validity of key components in Eq. \ref{eq:actorlossdasp}. We first generated two sets of actions for a given set of states from dataset: one set with safe outcomes, generated by a well trained policy in the medium-expert dataset; the other set with unsafe outcomes, composed of a series of random actions. We then utilized either the true dynamics model (TDM) or our DASP model to predict the next states of these actions and assess their safety as $score= \mathbb E_{s\sim D,a\sim\pi(\cdot|s)}\exp(\mathcal{R}(s,a)).$

\begin{table}[H] 
\centering
\caption{Validation study of DASP term.} 
\begin{tabular}{@{}lccc@{}} 
\toprule                  & Halfcheetah  & Hopper  & Walker2d  \\ \midrule 
TDM w. safe action        & 0.61 & 0.44 & 0.42 \\ 
TDM w. unsafe action      & 0.37 & 0.21 & 0.30 \\ \midrule
DASP w. safe action       & 0.64 & 0.47 & 0.44 \\ 
DASP w. unsafe action     & 0.38 & 0.19 & 0.27 \\ 
\bottomrule 
\end{tabular} \label{validityanalysis}
\end{table}

Table \ref{validityanalysis} shows the results. Comparing the results of the first and second rows, we observe that our safety score is sensitive to whether the consequences of actions are in-distribution (ID) or OOD, which supports the validity of this measurement. Analyzing the results from the third and fourth rows, we observe a notable score disparity in the density indicator between the two types of actions when utilizing the DASP model. This difference is similar to what we see in the first and second rows. It indicates that the DASP model performs well enough to differentiate between safe and unsafe actions.


\subsection{Ablation study}\label{sensitive_analysis1}

The DASP weight $\alpha$ is the hyperparameter that control the magnitude of how the DASP term influence the training. Its influence to DASP is as shown in Table \ref{betafigssb}, where three agents are all trained on the 'meidum' datasets. 
From the results, we note that the best choice for $\alpha$ in this implementation is around 0.1 for the "halfcheetah" and "hopper" tasks, while for the "walker2d" task, the optimal $\alpha$ is 0.05. We utilized these parameters in our experiments to achieve the best performance across the different tasks. 

\begin{table}[h]
\centering
\caption{The ablation study results of $\alpha$. We bold the highest scores in each task.}
\begin{tabular}{llll}
\toprule
$\alpha$ & Ha.-m & Ho.-m & Wa.-m \\
\midrule
0.01 & 66.0 & 104.8 & 101.6 \\
0.05 & 67.8 & 105.1 & \textbf{108.6} \\
0.1  & \textbf{70.4} & \textbf{108.6} & 100.0 \\
0.5  & 66.8 & 105.1 & 104.7 \\
3    & 65.0 & 104.3 & 102.5 \\
10   & 64.8 & 103.2 & 97.6  \\
100  & 51.6 & 100.8 & 85.6  \\
\bottomrule
\end{tabular}\label{betafigssb}
\end{table}

The results suggest that while moderate values of $\alpha$ enhance performance by balancing conservatism and generalization, excessive values lead to instability and poorer decision-making.

\textit{More experimental details, such as the structures of neural networks and the selection of hyperparameters, are available in Appendix \ref{externel_experiments}.}

\section{Conclusion}

In this paper, we propose a novel method called Density-Aware Safety Perception (DASP) to perform OOD state correction for a more robust and reliable offline reinforcement learning. To be specific, DASP is designed under a variational framework to achieve a more source-efficiency structure, which formulates the one-step forward dynamics model and the density model in a compact manner. Empirical results show that the proposed DASP outperforms most SOTA methods in offline RL, hence demonstrating the advantages of our method, which only uses an indicator instead of estimating specific distributions for OOD state correction.

\newpage

\bibliographystyle{named}
\bibliography{ijcai25}

\newpage

\appendix

\begin{LARGE}
\textbf{Appendix}
\end{LARGE}

\section{Proof of Theorem \ref{theorem:varia}.}\label{appendix:proof1}

\noindent\textbf{Theorem \ref{theorem:varia}.}
    \textit{The term $\mathbb E_{s'\sim P(s'|s,a)}\log d(s')$ could be lower bounded by solving the following optimization problem in the offline setting,}
\begin{align}
    &\max_{q_1,q_2}  \mathbb E_{(s,a,s')\sim\mathcal{D}} \bigg[\int dz\cdot q_1(z|s') \log P(s'|z)\nonumber \\
    &-KL(q_2(z|s,a)\| P(z))- KL(q_1(z|s')\|q_2(z|s,a)) \bigg]\label{eq:final_va_obj1}
\end{align}
    \textit{where $q_1(z|s')$ and $q_2(z|s,a)$ are two variational distributions. $KL(\cdot\|\cdot)$ is the KL-divergence between two distributions. $P(s'|z)$ is the poster distribution.}

\begin{proof}
    First, we introduce the first variational distribution $q_1(z|s')$
and by the Total Probability Equation $\int q_1(z|s') dz=1$, we have,
\begin{align}
    &\mathbb E_{s'\sim P(\cdot|s,a)}\log d(s') \label{eq:variational_obj1}\\
    = &\int ds' P(s'|s,a)\int dz\cdot q_1(z|s') \log d(s')\label{eq:81}
\end{align}
Then we introduce the second variational distribution $q_2(z|s,a)$, such that $\int P(s'|s,a)q_1(z|s')ds' = q_2(z|s,a)$, and the Bayes equation $d(s')=\frac{P(s',z)}{P(z|s')}$. So the above formulation in Eq.(\ref{eq:81}) could be transferred into,
\begin{align}
    & \int ds' P(s'|s,a)\int dz\cdot q_1(z|s') \log \frac{P(s',z)}{q_2(z|s,a)} \label{eq:deal11}\\
    &+ \int ds' P(s'|s,a)\int dz\cdot q_1(z|s') \log \frac{q_2(z|s,a)}{P(z|s')}\label{eq:deal21}
\end{align}

Here we deal with the term in Eq.(\ref{eq:deal21}) ,
\begin{align}
    Eq.(\ref{eq:deal21})& = \int ds' P(s'|s,a)KL(q_1(z|s')\|P(z|s')) \\
    &- \int ds' P(s'|s,a)KL(q_1(z|s')\|q_2(z|s,a))\\
    \geq & - \int ds' P(s'|s,a)KL(q_1(z|s')\|q_2(z|s,a))
\end{align}

Then we focus on the term in Eq.(\ref{eq:deal11}),
\begin{align}
    Eq.(\ref{eq:deal11})&=  \int dz\cdot q_2(z|s,a) \log \frac{P(z)}{q_2(z|s,a)}\\
    & +\int ds' P(s'|s,a)\int dz\cdot q_1(z|s') \log P(s'|z)\\
    &=  -KL(q_2(z|s,a)\| P(z))\\
    & +\int ds' P(s'|s,a)\int dz\cdot q_1(z|s') \log P(s'|z)
\end{align}
where the first equation is due to the aforementioned condition that $\int P(s'|s,a)q_1(z|s')ds' = q_2(z|s,a)$ - in practice, we will minimizing the KL-divergence between the two distributions $KL(\mathbb E_{P(s'|s,a)}q_1(z|s')\| q_2(z|s,a))$ to satisfy the equation, which would be discussed later.

To summary, the variational objective in Eq.(\ref{eq:oodcorrection1}) could be lower bounded by solving the following optimization problem in the offline setting,
\begin{align}
    &\max_{q_1,q_2}  \mathbb E_{(s,a,s')\sim\mathcal{D}} \bigg[\int dz\cdot q_1(z|s') \log P(s'|z)\nonumber \\
    &-KL(q_2(z|s,a)\| P(z))- KL(q_1(z|s')\|q_2(z|s,a)) \bigg]
\end{align}
\paragraph{Connection between two variational distributions.} In addition, it is worth noting that the problems of minimization of $KL(\mathbb E_{P(s'|s,a)}q_1(z|s')\| q_2(z|s,a))$ and $E_{P(s'|s,a)}KL(q_1(z|s')\|q_2(z|s,a))$ by according to $q_1,q_2$ may be redundant. That is, in some case, such as in the offline setting, the two optimization problems are equivalent, although the relationship between the two formulas is not obvious due to the nonlinear property of $KL$-divergence. However, in practice, we may utilize a Monte-Carlo approximation onto these terms, i.e., $KL(\frac{1}{N}\sum_{i=1}^Nq_1(z|s'_i)\| q_2(z|s,a))$ and $\frac{1}{N}\sum_{i=1}^N KL(q_1(z|s'_i)\|q_2(z|s,a))$. In offline setting, there is often $N=1$ for the lack of the dynamics model $P(s'|s,a)$. In this way, the two terms are both approximated with $\mathbb E_{(s,a,s')\sim\mathcal{D}}KL(q_1(z|s')\| q_2(z|s,a))$, so we only use this term in the variational result in Eq.(\ref{eq:final_va_obj1}), hence fulfill the gap.

Completing the proof.

\end{proof}

\section{External experiments}\label{externel_experiments}

\subsection{Code}\label{code}

 We build the proposed based on the RORL project from github\footnote{Project of RORL: https://github.com/YangRui2015/RORL}. The reasons why we choose YangRui2015's project are as follows: 1) The RORL framework is a classic baseline for the conservative offline reinforcement learning based on an implementation of PBRL~\cite{pbrl}. 2) Learning conservative Q functions can be easily implemented using the RORL framework. 3) To our knowledge, the RORL framework is the baseline with the highest scores in MuJoCo benchmarks. Our code is provided in the supplemental material. 

\subsection{Training details}\label{appendix:trainingprocedure}

In this section, we introduce our training details, including: 1) the hyperparameters our method use; 2) the structure of the neural networks we use: the Q-networks, inverse dynamics model network and policy network; 3) the training details of DASP; 4) the total amount of compute and the type of resources used.

\subsubsection{Hyperparameters of DASP}

In Table \ref{hpt3} and Table \ref{hpt4}, we give the hyperparameters used by DASP to generate Table \ref{comparison} and Table \ref{zaoshengshiyan} results. The $\alpha$ is the weight of the support-based constrain. 

\begin{table}[H]
\centering
\caption{Hyperparameters of DASP in standard MuJoCo benchmarks.}
\begin{tabular}{llll}
\toprule
\textbf{}         & \textbf{Halfcheetah} & \textbf{Hopper} & \textbf{Walker2d} \\ \midrule
$\alpha$          & 0.1                  & 0.1             & 0.05              \\
$\sigma$ & 0.001                        & 0.005                  & 0.01                     \\
  \bottomrule
\end{tabular}
\label{hpt3}
\end{table}

\begin{table}[H]
\centering
\caption{Hyperparameters of DASP in adversarial attack and OOS MuJoCo benchmarks.}
\begin{tabular}{llll}
\toprule
\textbf{}         & \textbf{Halfcheetah} & \textbf{Hopper} & \textbf{Walker2d} \\ \midrule
$\alpha$          & 0.1                  & 0.1             & 0.1              \\
$\sigma$ & 0.05                        & 0.005                  & 0.07                     \\
 \bottomrule
\end{tabular}
\label{hpt4}
\end{table}

\subsubsection{Neural network structures of DASP}

In this section, we introduce the structure of the networks we use in this paper: policy network, Q network and the dynamics model network. 

The structure of the policy network and Q networks is as shown in Table \ref{policyq}, where 's\_dim' is the dimension of states and 'a\_dim' is the dimension of actions. 'h\_dim' is the dimension of the hidden layers, which is usually 256 in our experiments. The policy network is a Guassian policy and the Q networks includes ten Q function networks and ten target Q function networks.

\begin{table}[h]
    \centering
    \caption{The structure of the policy net and the Q networks.}
    \begin{tabular}{ll}
        \toprule
policy net & Q net\\
		\midrule
Linear(s\_dim, h\_dim) & Linear(s\_dim + a\_dim, h\_dim)\\
Relu() & Relu()\\
Linear(h\_dim, h\_dim) & Linear(h\_dim, h\_dim)\\
Relu() & Relu()\\
Linear(h\_dim, a\_dim) & Linear(h\_dim, 1)\\
        \bottomrule
    \end{tabular}
    \label{policyq}
\end{table}

 The structure of the dynamics network is as shown in Table \ref{dmn}, which is a conditional variational auto-encoder. 's\_dim' is the dimension of states, 'a\_dim' is the dimension of actions and 'h\_dim' is the dimension of the hidden variables. 'z\_dim' is the dimension of the Gaussian hidden variables in conditional variational auto-encoder.

\begin{table}[h]
    \centering
    \caption{The structure of the density model network.}
    \begin{tabular}{l|l}
        \toprule
\multicolumn{2}{l}{density model net}\\
		\midrule
$q_1(z|s')$ & $q_2(z|s,a)$ \\
		\midrule
Linear(s\_dim, h\_dim) & Linear(s\_dim + a\_dim, h\_dim) \\
Relu() & Relu() \\
Linear(h\_dim, h\_dim)  & Linear(h\_dim, h\_dim) \\
Relu()  & Relu() \\
Linear(h\_dim, z\_dim) & Linear(h\_dim, z\_dim) \\
		\midrule
\multicolumn{2}{l}{$P(s'|z)$}  \\
		\midrule
\multicolumn{2}{l}{Linear(z\_dim, h\_dim)} \\
\multicolumn{2}{l}{Relu()}  \\
\multicolumn{2}{l}{Linear(h\_dim, h\_dim)}  \\
\multicolumn{2}{l}{Relu()}  \\
\multicolumn{2}{l}{Linear(h\_dim, s\_dim)} \\
        \bottomrule
    \end{tabular}
    \label{dmn}
\end{table}

\subsubsection{Training curves of DASP}
\begin{figure*}[htbp]
\centering
\includegraphics[width=0.9\linewidth]{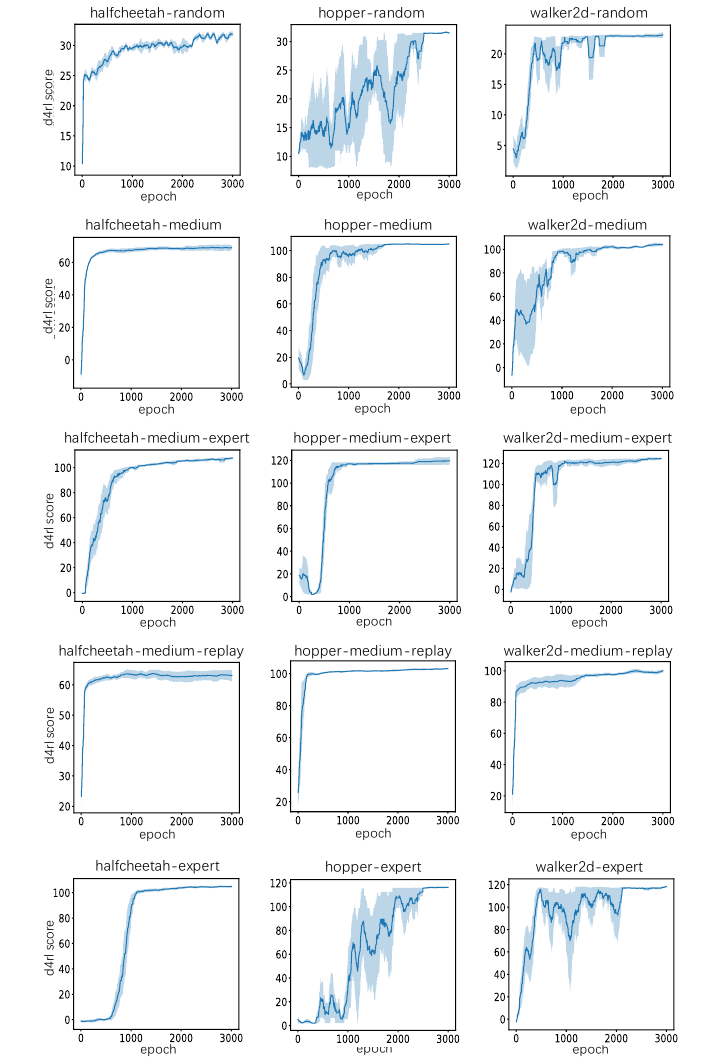}
\caption{Training curves of DASP on standard MuJoCo benchmarks over 4 seeds. One epoch corresponds to 1000 gradient steps.}
\label{traindetail}
\end{figure*}

We present the training curve of DASP from Table \ref{comparison} in Figure \ref{traindetail}. Each environment was trained for 3000 epochs, with each epoch corresponding to 1000 gradient steps.

\subsubsection{Compute resources}

We conducted all our experiments using a server equipped with one Intel Xeon Gold 5218 CPU, with 32 cores and 64 threads, and 256GB of DDR4 memory. We used a NVIDIA RTX3090 GPU with 24GB of memory for our deep learning experiments. All computations were performed using Python 3.8 and the PyTorch deep learning framework.

\section{Limitations}

\paragraph{Generalization boundary.} Just like the methods based on the traditional {\it state recovery} principle, the proposed DASP is also unable to generalize to those states that are quite far away from the offline dataset, where any action executed would not lead to any low-uncertainty state. In this situation, the DASP term would not embed any useful information for the new policy, because all the forward consequences have high uncertainty, which make such guidance degrade to a random-walk. Exploring the performance boundary of DASP is also a major direction for our future work.

\paragraph{Sensitivity to hyperparameters. } From the ablation study, we observe that the proposed method is sensitive to the selection of the hyperparameter weight coefficient $\alpha$. This problem can be alleviated by methods like Bayes optimization, which, however, is not the main research focus of this paper.


\end{document}